\documentclass[letterpaper]{article} %
\usepackage{aaai2026}  %
\usepackage{amsmath}
\usepackage{amssymb}
\usepackage{dsfont}
\usepackage{times}  %
\usepackage{helvet}  %
\usepackage{courier}  %
\usepackage[hyphens]{url}  %
\usepackage{graphicx} %
\usepackage{natbib}  %
\usepackage{caption} %
\usepackage{algorithm}
\usepackage{algorithmic}
\usepackage{booktabs}    %
\usepackage{multirow}    %
\usepackage{lscape}      %
\usepackage{xcolor}

\usepackage{newfloat}
\usepackage{listings}
\DeclareCaptionStyle{ruled}{labelfont=normalfont,labelsep=colon,strut=off} %
\floatstyle{ruled}
\newfloat{listing}{tb}{lst}{}
\floatname{listing}{Listing}
\title{Aligning Attention with Human Rationales for Self-Explaining Hate Speech Detection}

\author{
    Brage Eilertsen\textsuperscript{\rm 1} 
    Røskva Bjørgfinsdóttir\textsuperscript{\rm 1} 
    Francielle Vargas\textsuperscript{\rm 2}
    Ali Ramezani-Kebrya\textsuperscript{\rm 1,3,4}
}
\affiliations{
    \textsuperscript{\rm 1}University of Oslo\\
    \textsuperscript{\rm 2}University of São Paulo\\
    \textsuperscript{\rm 3}Integreat - Norwegian Centre for knowledge-driven machine learning\\
    \textsuperscript{\rm 4}TRUST – The Norwegian Centre for Trustworthy AI\\
    brageei@uio.no, roskvatb@uio.no, ali@uio.no 
}

\usepackage{bibentry}

\begin{document}
\maketitle
\begin{abstract}
The opaque nature of deep learning models presents significant challenges for the ethical deployment of hate speech detection systems. To address this limitation, we introduce Supervised Rational Attention (SRA), a framework that explicitly aligns model attention with human rationales, improving both interpretability and fairness in hate speech classification. SRA integrates a supervised attention mechanism into transformer-based classifiers, optimizing a joint objective that combines standard classification loss with an alignment loss term that minimizes the discrepancy between attention weights and human-annotated rationales.
We evaluated SRA on hate speech benchmarks in English (HateXplain) and Portuguese (HateBRXplain) with rationale annotations. Empirically,  SRA achieves 2.4× better explainability compared to current baselines, and produces token-level explanations that are more faithful and human-aligned. In terms of fairness, SRA achieves competitive fairness across all measures, with second-best performance in detecting toxic posts targeting identity groups, while maintaining comparable results on other metrics. These findings demonstrate that incorporating human rationales into attention mechanisms can enhance interpretability and faithfulness without compromising fairness.

\end{abstract}

\section{Introduction}
The proliferation of social media platforms has necessitated the development of automated hate speech detection systems. These systems operate at large scale, with platforms such as Meta processing over seven million hate speech appeals monthly \cite{gorwa2020algorithmic,meta2025oversight}. These automated classifiers pose the risk of reinforcing societal biases and marginalizing vulnerable groups unless carefully designed and evaluated for fairness and representational harms \cite{davidson-etal-2019-racial,davani2023,ungless-etal-2025-way,attanasio-etal-2022-entropy,vargas-etal-2023-socially}.
In this work, we define offensive language as confrontational, rude, or aggressive content \cite{Davidson_Warmsley_Macy_Weber_2017,zampierietal2019}, while hate speech specifically targets individuals based on their social identities \cite{FortunaAndNunes2018}. Sentiments can be expressed explicitly or implicitly \cite{Polettoetall2021,Vargas_Carvalho_Pardo_Benevenuto_2024}. Existing automated classifiers exhibit systematic biases such as oversensitivity to identity terms, reinforcement of unfair associations of ethnic stereotypes, favouring European American names over African American names and associating negative sentiments with people with disabilities \cite{doi:10.1126/science.aal4230,hutchinson-etal-2020-social,10.1145/3278721.3278729}.

While deep learning (DL) approaches achieve high performance in hate speech detection, they present two critical limitations. First, their black-box nature prevents understanding of whether decisions rely on problematic shortcuts or genuine indicators of harm \cite{gongane2024survey}. Second, systematic biases in hate speech detection can marginalize the communities they aim to protect \cite{may-etal-2019-measuring}.
Developing transparent and responsible AI systems has essential important due to both ethical considerations and regulatory requirements \cite{8808038,gdpr2016}. Given the potential consequences of misidentifying or overlooking harmful content, developing interpretable hate speech detection systems represents a key challenge in Natural Language Processing (NLP) research \cite{mathew2021hatexplain,calabrese-etal-2024-explainability,salles-etal-2025-hatebrxplain}. Explainable AI methods that clarify the rationales (supporting evidence) behind predictions are the key components for the ethical deployment of such systems \cite{balkir-etal-2022-challenges}.

Prior work has explored aligning model attention with human rationales, text snippets that guide and justify labeling decisions \cite{deyoung-etal-2020-eraser,jain-etal-2020-learning}. Rationale-based learning improves interpretability across NLP tasks including Natural Language Inference, Question Answering, Information Retrieval, and Sentiment Analysis \cite{deyoung-etal-2020-eraser,bastings-etal-2019-interpretable,lei-etal-2016-rationalizing,zhang-etal-2016-rationale,jiang-etal-2021-alignment,lehman-etal-2019-inferring,jorgensen-etal-2022-multilingual}. In these applications, attention-based models \cite{10.5555/3295222.3295349} have been explored since they offer a certain degree of interpretability. However, these explanations are often superficial and lack insights into why a model might consider specific features or tokens as relevant. For example, \citet{strout-etal-2019-human} found that explanations generated using supervised attention are judged superior compared to explanations generated using normal unsupervised attention. \citet{bao-etal-2018-deriving} also showed that mapping human-annotated rationales into a continuous space significantly improves over the baselines and reduces error by over 15\% on average across benchmark datasets. These studies did not, however, address fairness implications or evaluate token-level explanations for sensitive applications.

We propose Supervised Rational Attention (SRA), a method that enhances the  explainability and fairness of hate speech detection systems. We argue that by using rationales that are explicitly labeled by domain experts, the system will generate more transparent and meaningful explanations. This can build trust with end-users and enable better decision-making in sensitive applications, such as content moderation.

Our contributions are summarized as follows:
\begin{itemize}
    \item We introduce SRA, a framework to align model attention with human-annotated rationales.

    \item We evaluate SRA on English (HateXplain) and Portuguese (HateBRXplain) benchmarks   \citep{mathew2021hatexplain, salles-etal-2025-hatebrxplain} and find that SRA improves explainability metrics (IoU F1 and Token F1) and achieves competitive fairness with second-best (GMB-BNSP) performance among all methods, with minimal impact on predictive performance.

    \item We provide publicly available code, datasets and models including rationale mask construction and attention supervision, to facilitate future research on trustworthy NLP and DL.
    
\end{itemize}

\section{Related Work}
Explainable hate speech detection methods are commonly categorized as either \textit{self-explaining} or \textit{post-hoc explaining} \cite{10.1145/3236009}. Self-explaining methods integrate interpretability directly into the model architecture, while post-hoc methods generate explanations after training using the model's input-output behavior \cite{balkir-etal-2022-challenges}. Our work focuses on self-explaining methods that leverage human rationales during training.

\paragraph{HateXplain Baseline.} 
\citet{mathew2021hatexplain} proposed the foundational approach for explainable hate speech detection with human rationales. They transform human-annotated text spans into ground truth attention vectors by averaging across annotators and applying softmax with temperature. During training, attention-based models (BiRNN-Attention and BERT) minimize cross-entropy loss between predicted attention weights and ground truth attention vectors, encouraging alignment with human rationales. Models are evaluated on both classification metrics (accuracy, macro-F1, AUROC) and explainability metrics following the ERASER benchmark \cite{deyoung-etal-2020-eraser}, including plausibility (IoU-F1, Token-F1, AUPRC) and faithfulness (comprehensiveness, sufficiency).

\paragraph{Attention Supervision Methods.} 
\citet{kim-etal-2022-hate} introduced Masked Rationale Prediction (MRP), which masks portions of rationale embeddings during an intermediate token-level classification task. The model learns to predict masked rationale labels before fine-tuning on hate speech classification. \citet{clarke-etal-2023-rule} developed
Rule by Example, a contrastive learning framework that grounds hate-speech predictions in logical rules rather than token-level rationales.

\paragraph{Limitations of Prior Work.} 
Existing attention supervision methods have improved explanation quality \cite{strout-etal-2019-human,bao-etal-2018-deriving}, with significant error reductions on benchmark datasets. However, these studies have not adequately addressed fairness implications or evaluated token-level explanations specifically for sensitive applications like hate speech detection, where both interpretability and equitable treatment across demographic groups are critical. Additional related work on  post-hoc and self-explaining methods is provided in Supplementary.

\section{Supervised Rational Attention (SRA)}
The SRA framework aligns model attention with human rationales for hate speech detection.
SRA enhances standard transformer-based text classifiers by explicitly aligning model attention with human-annotated rationales, leading to explanations that are plausible and faithful~\citep{deyoung-etal-2020-eraser}.

\paragraph{Problem Setting.}
Let $x = (w_1, \ldots, w_L)$ denote a tokenized input sequence, and $y \in \{0, 1, \ldots, C-1\}$ its class label, where $C$ is the number of classes (e.g., $C=3$ for multiclass hate speech classification). In the HateXplain benchmark~\citep{mathew2021hatexplain}, labels are defined as normal~$=0$, offensive~$=1$, and hate speech~$=2$. For a subset of training examples, we are additionally given a \textit{rationale mask} $r = (r_1, \ldots, r_L)$, where $r_i \in \{0,1\}$ indicates whether token $w_i$ is part of the human rationale for the label $y$.

\paragraph{Model Architecture.}
We use a pre-trained transformer encoder $f_\theta$ (e.g., BERT or BERTimbau) that computes contextual representations $\mathbf{h}_i$ for each input token $w_i$. A classification head predicts the class label:
\[
\hat{y} = \arg\max_{c \in \{0, 1, ..., C-1\}} \operatorname{softmax}(\mathbf{W}_c \mathbf{h}_{\text{[CLS]}} + \mathbf{b}_c)
\]
where $\mathbf{h}_{\text{[CLS]}} \in \mathbb{R}^d$ is the representation of the [CLS] token, $\mathbf{W}_c \in \mathbb{R}^{1 \times d}$ and $\mathbf{b}_c \in \mathbb{R}$ are the learned weight matrix and bias term for class $c$, and $C$ is the number of classes ($C=3$ for HateXplain, $C=2$ for HateBRXplain).

\paragraph{Attention Extraction.}
Let $A^{(l,h)} \in [0,1]^{L \times L}$ denote the attention matrix at layer $l$ and head $h$ of encoder $f_\theta$. To obtain model explanations, we first extract the attention from the [CLS] token to all input tokens:
\[
\mathbf{a}^{(l,h)} = (a^{(l,h)}_1, \ldots, a^{(l,h)}_L) = A^{(l,h)}_{\text{[CLS]}, :} \in \mathbb{R}^L
\]
To select appropriate attention weights for supervision, we have conducted comprehensive ablation studies across layers and heads (see Supplementary for detailed results). Our experiments show that SRA achieves consistent performance across layers 6-11 and all attention heads, with minimal variation in both classification performance and explainability metrics. This robustness indicates that our method is not dependent on specific architectural choices. Based on these ablation studies, we use layer 8 and head 7 for our main experiments, as this configuration provides a good balance between classification and explainability performance. %
To reduce  computational costs, we also evaluated a variant of SRA by taking an average of attention weights over all heads within a layer. Our experiments for this variant show consistent results with details provided in Supplementary.   

\paragraph{Attention Alignment Loss (AAL).}
Given a sample with rationale mask $r$, we encourage the model to attend to tokens deemed important by human annotators by minimizing the mean squared error (MSE) between normalized attention $\mathbf{a}$ and $r$:
\[
\ell_{\mathrm{AAL}}(\mathbf{a}, r) = \frac{1}{\sum_{i=1}^L m_i} \sum_{i=1}^L m_i \left( \frac{a_i}{\sum_{j=1}^L m_j a_j + \epsilon} - r_i \right)^2
\]
where $m_i \in \{0,1\}$ is the padding mask (1 for valid tokens, 0 for padding) and $\epsilon = 10^{-10}$ for numerical stability.

\paragraph{Overall Training Objective.}
We combine the standard cross-entropy loss for classification and the supervised attention alignment loss:
\[
\ell_{\mathrm{total}} = \ell_{\mathrm{CE}}(\hat{y}, y) + \alpha \mathbf{1}[y > 0] \mathbf{1}\left[\sum_{i=1}^L r_i > 0\right] \ell_{\mathrm{AAL}}(\mathbf{a}, r)
\]
where $\alpha$ is a hyperparameter controlling the strength of attention supervision, balancing attention alignment and classification, $\mathbf{1}[y > 0]$ is an indicator function that equals 1 when the label is offensive or hate speech (i.e., $y \in \{1,2\}$), and $\mathbf{1}[\sum_{i=1}^L r_i > 0]$ ensures rationales exist. The supervised attention loss is only applied for training examples labeled as offensive or hate speech with human-annotated rationales.

\paragraph{Rationale Extraction (Dataset-dependent).}
\begin{itemize}
    \item For datasets with \textit{token-level rationales}, $r$ is obtained by majority voting \textbf{across annotators who provided rationales for each example}. Specifically, a token is included in the rationale mask if it is selected by at least 50\% of the available annotators (i.e., those who supplied rationales), corresponding to a threshold of 0.5.
    \item For datasets with \textit{free-text rationale spans}, $r$ is constructed by mapping character-level rationale spans to token indices using offset mappings from the tokenizer.
\end{itemize}
All rationale masks are aligned with the tokenization of $x = (w_1, \ldots, w_L)$ and truncated or padded to the model's maximum input length.

\paragraph{Inference and Explainability.}
At inference, the same attention mechanism provides explanation for each prediction: the tokens with highest attention from [CLS] correspond to the model's ``rationale'' for its decision. This allows evaluation of faithfulness and plausibility metrics~\citep{deyoung-etal-2020-eraser} with respect to human rationales.

\section{Experimental Setup}

\paragraph{Implementation Overview.}
We implement our framework in PyTorch with Hugging Face Transformers. Full code, including rationale mask construction, will be released upon publication.

\paragraph{Models.}
We use \textbf{BERT-base-uncased}~\cite{devlin-etal-2019-bert} for English and \textbf{BERTimbau-base}~\cite{souza2020bertimbau} for Portuguese, extracting attention from layer 8 and head 7 based on preliminary validation (see Table~\ref{tab:ablation_layers} in Supplementary for Ablation study).

\paragraph{Datasets.}
Models are trained on \textbf{HateXplain}~\cite{mathew2021hatexplain} (English, 20,148 samples with three classes; strong inter-annotator reliability) and \textbf{HateBRXplain}~\cite{Vargas_Carvalho_Pardo_Benevenuto_2024} (Portuguese, 7,000 samples with two classes). We created 80/10/10 stratified train/validation/test splits with no data overlap between splits, using fixed random seeds for reproducibility, and converted rationales to binary masks via majority vote across annotators.

\paragraph{Training.}
Hyperparameters are tuned on validation macro F1. We use AdamW (English: $2 \times 10^{-5}$, batch 16, max len 128; Portuguese: $1 \times 10^{-5}$, batch 8, max len 512), 5 epochs, and select the checkpoint with best validation F1. Unless otherwise specified, we set $\alpha=10.0$ in the overall objective. All results are averaged over 5 random seeds.

\paragraph{Hardware.}
Our experiments have been conducted on a cluster node with one A100 GPU. %

\section{Results and Discussion}
Our results show that explicit rationale supervision can improve explainability (2.4× better IoU F1) while maintaining classification performance and competitive fairness across demographic groups.

We compare SRA with the following baselines from~\cite{mathew2021hatexplain} on HateXplain (English):
(1) \textbf{CNN-GRU}~\cite{zhang2018hate}, which combines CNN filters with GRU layers;
(2) \textbf{BiRNN}~\cite{schuster1997bidirectional}, a bidirectional RNN model;
(3) \textbf{BiRNN-Attention}~\cite{liu2016attention}, which adds an attention mechanism to BiRNN;
and (4) \textbf{BERT}~\cite{devlin-etal-2019-bert}, the transformer-based model.
For models with attention mechanisms (BiRNN-Attention and BERT), we evaluate both unsupervised attention variants (denoted with suffix ``-Attn'') and variants trained with attention supervision (denoted with suffix ``-HateXplain''). Additionally, we compare against BERT-MRP~\cite{kim-etal-2022-hate}, a masked rationale prediction approach, and post-hoc explanation methods [LIME]~\citep{ribeiro2016lime} on the baseline models. The bracketed labels [Attn] and [LIME] indicate the explanation method used: attention-based or LIME post-hoc explanations, respectively. Table~\ref{tab:main_results} uses these labels for clarity.

We evaluate SRA on both English and the Portuguese benchmarks, showing consistent improvements in attention alignment, explainability metrics, and bias reduction across languages.

\begin{table*}[t]
\centering
\caption{Performance comparison of hate speech detection models on the HateXplain test set. We evaluate classification performance (Accuracy, Macro F1, AUROC), explainability metrics (IoU F1, Token F1, AUPRC), fairness metrics (GMB-Subgroup, GMB-BPSN, GMB-BNSP AUCs), and faithfulness metrics (Comprehensiveness, Sufficiency). Higher values are better for all metrics except Sufficiency (lower is better). Models with [LIME] use post-hoc explanations, while [Attn] indicates attention-based explanations. Our proposed SRA method achieves the best IoU F1 and Token F1 explainability scores while maintaining competitive classification performance, fairness, and faithfulness. Best results are in \textbf{bold}, second-best are \underline{underlined}. SRA results are averaged across 5 random seeds with standard deviations shown in parentheses.}
\label{tab:main_results}
\resizebox{\textwidth}{!}{
\begin{tabular}{l|ccc|ccc|ccc|cc}
\toprule
\multirow{2}{*}{\textbf{Model [Explanation Method]}} 
    & \multicolumn{3}{c|}{\textbf{Classification}} 
    & \multicolumn{3}{c|}{\textbf{Explainability}} 
    & \multicolumn{3}{c|}{\textbf{Bias (AUC)}} 
    & \multicolumn{2}{c}{\textbf{Faithfulness}} \\
\cmidrule(lr){2-4} \cmidrule(lr){5-7} \cmidrule(lr){8-10} \cmidrule(lr){11-12}
& Acc.$\uparrow$ & Macro F1$\uparrow$ & AUROC$\uparrow$
& IoU F1$\uparrow$ & Token F1$\uparrow$ & AUPRC$\uparrow$
& GMB-Sub.$\uparrow$ & GMB-BPSN$\uparrow$ & GMB-BNSP$\uparrow$
& Comp.$\uparrow$ & Suff.$\downarrow$ \\
\midrule
CNN-GRU [LIME] & 0.627 & 0.606 & 0.793 & 0.167 & 0.385 & 0.648 & 0.654 & 0.623 & 0.659 & 0.316 & \textbf{-0.082} \\
BiRNN [LIME] & 0.595 & 0.575 & 0.767 & 0.162 & 0.361 & 0.605 & 0.660 & 0.640 & 0.671 & 0.421 & -0.051 \\
BiRNN-Attn [Attn] & 0.621 & 0.614 & 0.795 & 0.167 & 0.369 & 0.643 & 0.653 & 0.662 & 0.668 & 0.278 & -0.001 \\
BiRNN-Attn [LIME] & 0.621 & 0.614 & 0.795 & 0.162 & 0.386 & 0.650 & 0.653 & 0.662 & 0.668 & 0.308 & \underline{-0.075} \\
BiRNN-HateXplain [Attn] & 0.629 & 0.629 & 0.805 & \underline{0.222} & \underline{0.506} & \textbf{0.841} & 0.691 & 0.691 & 0.674 & 0.281 & 0.039 \\
BiRNN-HateXplain [LIME] & 0.629 & 0.629 & 0.805 & 0.174 & 0.407 & 0.685 & 0.691 & 0.691 & 0.674 & 0.343 & \underline{-0.075} \\
BERT [Attn] & 0.690 & 0.674 & 0.843 & 0.130 & 0.497 & \underline{0.778} & 0.762 & 0.709 & 0.757 & 0.447 & 0.057 \\
BERT [LIME] & 0.690 & 0.674 & 0.843 & 0.118 & 0.468 & 0.747 & 0.762 & 0.709 & 0.757  & 0.436 & 0.008 \\
BERT-HateXplain [Attn] & \underline{0.698} & \underline{0.687} & 0.851 & 0.120 & 0.411 & 0.626 & \underline{0.807} & \underline{0.745} & 0.763 & 0.424 & 0.160 \\
BERT-HateXplain [LIME] & \underline{0.698} & \underline{0.687} & 0.851 & 0.112 & 0.452 & 0.722 & \underline{0.807} & \underline{0.745} & 0.763 & \textbf{0.500} & 0.004 \\
BERT-MRP [Attn] & \textbf{0.704} & \textbf{0.699} & \textbf{0.862} & 0.141 & 0.504 & 0.745 & \textbf{0.815} & \textbf{0.748} & \textbf{0.854} & \underline{0.479} & 0.067 \\

\midrule
\textbf{SRA, Ours ($\alpha=10$)} 
    & 0.696 %
    & 0.682 %
    & \underline{0.855} %
    & \textbf{0.539} %
    & \textbf{0.651} %
    & 0.753 %
    & 0.714 %
    & 0.718 %
    & \underline{0.835} %
    & 0.417 %
    & -0.013 %
\\
& (±0.007) & (±0.010) & (±0.002) & (±0.005) & (±0.002) & (±0.001) & (±0.002) & (±0.003) & (±0.012) & (±0.019) & (±0.012) \\
\bottomrule
\end{tabular}}
\end{table*}

\subsection{Results on HateXplain (English)  in~Table~\ref{tab:main_results}}

\paragraph{Classification Performance.}
SRA achieves a macro F1 score of 0.682 and accuracy of 0.696, marginally below the best-performing BERT-MRP baseline (F1: 0.699, Acc: 0.704). SRA achieves AUROC of 0.855 that is competitive with rationale-based baselines, including BERT-MRP (0.862) and BERT-HateXplain (0.851). This small performance difference (less than 2\% in F1) suggests that incorporating human rationale supervision can significantly improve interpretability with little impact on classification performance.

\paragraph{Explainability Improvements.}
SRA shows improvements in explainability metrics compared to the baselines. \textbf{IoU F1}, which measures the Intersection-over-Union F1 score between model attention and human rationales (with matches defined by an overlap exceeding 0.5 \cite{deyoung-etal-2020-eraser}), improves to 0.539—representing a 4.5× improvement over BERT-HateXplain's supervised attention and a 2.4× gain over the best baseline (BiRNN-HateXplain). \textbf{Token F1}, which evaluates rationale alignment as token-level binary classification based on precision and recall for individual token matches \cite{deyoung-etal-2020-eraser}, reaches 0.651, outperforming the best baseline by 29\%. For SRA, token-level precision and recall are 0.937 and 0.579, respectively, indicating that when the model highlights tokens, they are likely to align with human rationales. The \textbf{AUPRC} metric evaluates soft token scoring by sweeping thresholds over attention weights and confirms SRA's strong ranking ability for rationale tokens.

These improvements indicate that SRA learns to focus on the same textual evidence that human annotators identify as indicators of hate speech.

\paragraph{Fairness Metrics.}
We evaluate fairness using three Generalized Mean Bias (GMB) metrics \cite{borkan-etal-2019-nuanced}. SRA achieves a \textbf{GMB-Subgroup AUC} of 0.714, which measures the model's ability to distinguish toxic from normal posts that mention identity groups. While this is lower than BERT-MRP's 0.815 and BERT-HateXplain's 0.807, it remains comparable. The \textbf{GMB-BPSN AUC} of 0.718 (compared to 0.748 for BERT-MRP and 0.745 for BERT-HateXplain) evaluates false-positive rates by measuring performance on normal posts mentioning identities versus toxic posts without identity terms. SRA achieves a \textbf{GMB-BNSP AUC} of 0.835, the second-best result after BERT-MRP's 0.854, showing improved performance at avoiding false negatives when toxic posts mention identity groups compared to BERT-HateXplain's 0.763. These results suggest a trade-off among fairness metrics. SRA shows slightly reduced performance on subgroup classification and false positive measures but significantly improves false negative detection for posts targeting identity groups compared to non-MRP baselines. This indicates that supervised rationale alignment helps the model better identify genuinely harmful content targeting protected groups.%

\paragraph{Faithfulness Analysis.}
Faithfulness metrics evaluate whether model explanations correspond to features actually used for predictions \cite{deyoung-etal-2020-eraser}. SRA achieves a \textbf{comprehensiveness} score of 0.417, which measures the performance drop when removing high-attention (rationale) tokens. Higher values indicate that rationales were influential in predictions. While SRA is outperformed by baselines using post-hoc explanation, e.g., BERT-HateXplain [LIME], it is competitive or outperforms other attention-based baselines. This suggests that the model uses both highlighted rationales and wider context. The \textbf{sufficiency} score compares the model's prediction confidence on the full input text versus its confidence when using only the rationale tokens. The negative values, which are common among attention-based methods including SRA, indicate that the model becomes more confident for prediction of the class given only rationales since it is less distracted by the context.

\begin{figure}[h]
    \centering
    \includegraphics[width=\linewidth]{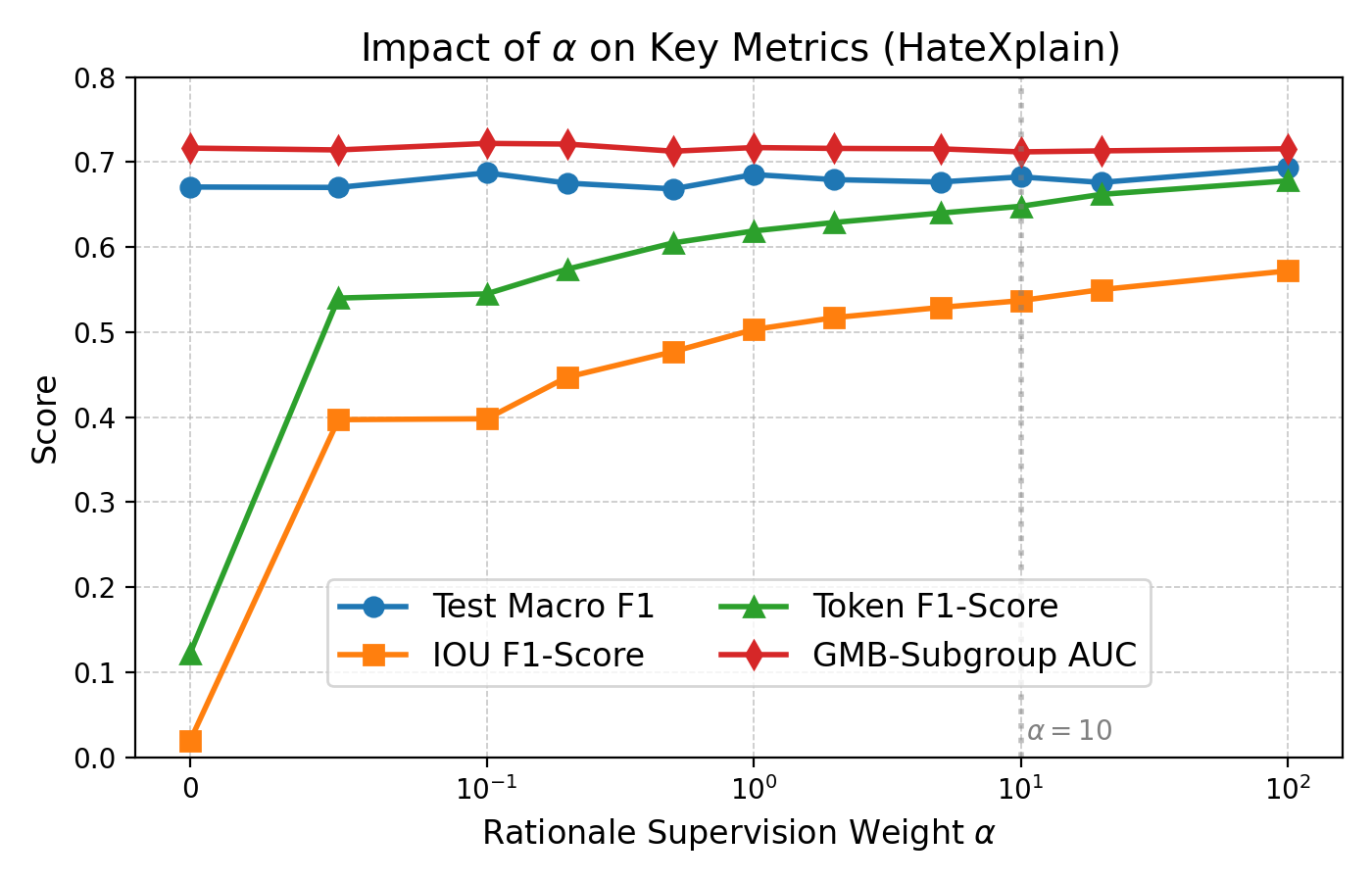}
    \caption{Impact of the rationale alignment hyperparameter} $\alpha$ on test performance, explainability, and fairness metrics for HateXplain dataset. Increasing $\alpha$ yields improvements in rationale alignment (IoU F1, Token F1) while maintaining stable classification performance (Test Macro F1) and fairness (GMB-Subgroup AUC). The vertical line indicates our chosen operating point ($\alpha=10$).
    \label{fig:alpha_ablation}
\end{figure}

\paragraph{Ablation and Robustness.}
To evaluate the robustness of our method to the choice of rationale alignment hyperparameter, we conduct an ablation study by varying $\alpha$ from $0$ (baseline without supervision) to $100$. As illustrated in Figure~\ref{fig:alpha_ablation}, increasing $\alpha$ improves explainability metrics: IoU F1 improves from 0.019 (baseline) to 0.572 ($\alpha=100$), while Token F1 increases from 0.122 to 0.678. These gains come with limited impact on classification performance. Test Macro F1 remains relatively stable across all $\alpha$ values. Similarly, fairness remains relatively stable, with GMB-Subgroup AUC ranging from 0.7112-0.7220. We use $\alpha=10$ as our primary configuration, achieving good explainability (IoU F1: 0.537, Token F1: 0.648) while maintaining comparable performance (Test F1: 0.683) and fairness (GMB-Sub: 0.7120). The range of effective $\alpha$ values suggests that SRA can substantially improve interpretability with minimal impact on performance and biases.%

\subsection{Results on HateBRXplain (Portuguese)}

\begin{table*}[t]
\centering
\caption{Comparison of explainability methods on the HateBRXplain (Portuguese) test set. We evaluate post-hoc explanation methods (LIME, SHAP) against our intrinsic SRA approach across multiple Portuguese language models. Explainability metrics (IoU F1, Token Precision/Recall/F1) measure alignment with human rationales, while faithfulness metrics (Comprehensiveness, Sufficiency) assess whether explanations reflect actual model reasoning. Higher values indicate better performance for all metrics except Sufficiency (lower is better). SRA provides explanations intrinsically during prediction, while LIME and SHAP require additional post-hoc computation. Our SRA method achieves superior token precision (0.935) and competitive overall performance while providing real-time explanations. Best results are in \textbf{bold}. SRA results are averaged across 5 random seeds with standard deviations shown in parentheses.}
\label{tab:portuguese_xai}
\scalebox{0.95}{
\begin{tabular}{l|ccccc|cc}
\toprule
\multirow{2}{*}{Model [XAI method]} & \multicolumn{5}{c|}{Explainability} & \multicolumn{2}{c}{Faithfulness} \\
& IoU F1$\uparrow$ & Token Prec$\uparrow$ & Token Rec$\uparrow$ & Token F1$\uparrow$ &  & Comp.$\uparrow$ & Suff.$\downarrow$ \\
\midrule
mBERT [LIME]        & 0.5828 & 0.7458 & 0.6936 & 0.6701 &  & 0.8809 & 0.0134 \\
mBERT [SHAP]        & 0.6628 & 0.7143 & 0.7520 & 0.6897 &  & 0.9324 & 0.0172 \\
BERTimbau [LIME]    & 0.5857 & 0.7557 & 0.6848 & 0.6698 &  & 0.9094 & 0.0237 \\
BERTimbau [SHAP]    & 0.6600 & 0.7489 & 0.7099 & 0.6831 &  & 0.8458 & 0.0215 \\
DistilBERTimbau [LIME] & 0.6457 & 0.7614 & 0.7276 & 0.7003 &  & 0.9407 & 0.0115 \\
DistilBERTimbau [SHAP] & 0.6200 & 0.7543 & 0.6862 & 0.6720 &  & \textbf{0.9475} & 0.0114 \\
PTTS [LIME]         & 0.6057 & 0.7487 & 0.6978 & 0.6776 &  & 0.5654 & 0.0016 \\
PTTS [SHAP]         & \textbf{0.7400} & 0.7177 & \textbf{0.8378} & 0.7362 &  & 0.6160 & 0.0083 \\
\midrule
\textbf{SRA, Ours ($\alpha=10$)} & 0.716 & \textbf{0.935} & 0.668 & \textbf{0.745} & & 0.454 & \textbf{-0.036} \\
& (±0.025) & (±0.005) & (±0.014) & (±0.010) & & (±0.114) & (±0.016) \\
\bottomrule
\end{tabular}
}
\end{table*}

For HateBRXplain, we compare against the following baselines from~\cite{salles-etal-2025-hatebrxplain}: (1)
mBERT (Devlin et al. 2019), the multilingual BERT model;
(2) BERTimbau (Souza, Nogueira, and Lotufo 2020),
a Portuguese-specific BERT model; (3) DistilBERTim-
bau (Junior 2024), a distilled version of BERTimbau; and
(4) PTT5 (Carmo et al. 2020), a Portuguese T5 model. Since
these baselines were evaluated using post-hoc explanation
methods, we compare against both LIME (Ribeiro, Singh,
and Guestrin 2016) and SHAP (Lundberg and Lee 2017) explanations for each model. The bracketed labels [LIME] and [SHAP] indicate the post-hoc explanation method used for each baseline model. %

\paragraph{Comparison with Post-hoc Methods.}
We compare SRA against LIME and SHAP explanations across multiple Portuguese language models in~Table~\ref{tab:portuguese_xai}. While post-hoc methods like PTTS [SHAP] achieve comparable IoU F1 scores, SRA with BERTimbau ($\alpha=10$) shows high token-level precision (0.935) and improved overall token F1 (0.745). SRA provides these explanations intrinsically during prediction, unlike post-hoc methods that require additional computation and may not reflect the model's actual decision process. This intrinsic explainability  makes SRA more appealing for real-time applications where computational efficiency is important.

\paragraph{Classification Performance.}
On HateBRXplain, SRA shows robust performance across varying rationale alignment hyperparameter values. Our ablation study (Table~\ref{tab:portuguese_ablation} in  Supplementary) reveals that Test Macro F1 remains stable across all values of $\alpha$, from baseline 0.903 to a peak of 0.921 at $\alpha=0.5$, with only minor variations (±2\%) even at $\alpha=100$. This stability, mirroring the pattern observed in English, suggests that incorporating rationale supervision does not compromise the model's hate speech detection capabilities. At $\alpha=10$, SRA achieves a Test F1 of 0.910 (±0.008) with an accuracy of 0.907 (±0.007) and AUROC of 0.966 (±0.004), demonstrating consistent performance.

\paragraph{Explainability Improvements.}
Similar to our English results, incorporating attention alignment loss yields monotonic improvements in explainability metrics (see Tables~\ref{tab:english_ablation} and~\ref{tab:portuguese_ablation} in Supplementary). IoU F1 increases from 0.387 at $\alpha=0.05$ to 0.751 at $\alpha=100$, while the baseline (no rationale alignment) achieves 0. Token F1 shows parallel gains, improving from 0.574 to 0.771. Token Precision peaks at $\alpha=1.0$, maintaining values above 0.91 across all supervised settings. At our selected operating point ($\alpha=10$), SRA achieves IoU F1 of 0.716 (±0.025), Token F1 of 0.745 (±0.010), with Token Precision of 0.935 (±0.005) and Token Recall of 0.668 (±0.014). These improvements suggest that SRA learns to focus on textual evidence that human annotators identify as critical for hate speech classification.

\paragraph{Fairness and Cross-lingual Validation.}
The ablation analysis shows that fairness metrics remain relatively stable across different values of $\alpha$. Detailed fairness results for both English and Portuguese datasets are provided in Tables~\ref{tab:bias_ablation} and \ref{tab:bias_ablation_pt} of Supplementary. GMB-Subgroup AUC varies slightly from 0.7165 (baseline) to 0.7157 ($\alpha=100$), indicating minimal trade-off between explainability and subgroup fairness. At $\alpha=10$, GMB-Subgroup AUC is 0.7120 (±0.003), GMB-BPSN AUC is 0.7211 (±0.002), and GMB-BNSP AUC improves to 0.8186 (±0.004), representing slight improvement over the baseline. These results suggest that supervised rationale alignment preserves fairness while improving explainability, with the strongest gains observed in reducing false negatives on identity-targeted toxic content (GMB-BNSP).
Similar patterns across HateXplain (English) and HateBRXplain (Portuguese) suggest SRA generalizes across languages. Despite differences in dataset characteristics, HateXplain with 20,148 multi-platform samples versus HateBRXplain with 7,000 Instagram-focused samples, both exhibit similar behaviors with stable classification performance across $\alpha$ values, substantial explainability improvements, and comparable fairness trade-offs. These results suggest SRA is applicable to interpretable hate speech detection in different languages.

\subsection{Qualitative Analysis}
While quantitative metrics show SRA's overall performance, qualitative examination of attention patterns provides insights into how rationale supervision influences model behavior in practice. Understanding these patterns is essential for hate speech detection systems, where explainability requirements demand that models not only achieve high performance, but also focus on linguistically meaningful features that align with human reasoning. We examine specific examples to illustrate how SRA addresses key challenges in self-explaining hate speech detection.

\begin{figure*}[t]
\centering
\small
\begin{tabular}{l|p{0.40\textwidth}|p{0.40\textwidth}}
\toprule
& \textbf{Baseline BERT ($\alpha=0$)} & \textbf{SRA ($\alpha=10$)} \\
\midrule
\textbf{Refugee Example} \\ Label: Hate speech & 
\colorbox{red!70}{allowing} \colorbox{red!37}{refugees} into \colorbox{red!36}{your} \colorbox{red!80}{nation} is like allowing rabid foxes into your chicken coop it does not make you caring it makes you an asshole &
allowing \colorbox{red!29}{refugees} into your nation is like allowing \colorbox{red!90}{rabid} \colorbox{red!65}{foxes} into your chicken coop it does not make you caring it makes you an \colorbox{red!27}{asshole} \\
& Human rationale: \textit{allowing, refugees, rabid, foxes, nation} & Human rationale: \textit{allowing, refugees, rabid, foxes, nation} \\
& \textit{Probabilities:} Normal: 2.59\%, Offensive: 90.17\%, Hate speech: 7.24\% & \textit{Probabilities:} Normal: 1.70\%, Offensive: 59.56\%, Hate speech: 38.74\% \\
\bottomrule
\end{tabular}
\caption{Attention heatmap visualization comparing baseline BERT ($\alpha=0$) and SRA ($\alpha=10$) on a test example involving implicit bias through metaphorical dehumanization. Color intensity represents attention weights from layer 8, head 7 (darker = higher attention). Baseline attention scatters across neutral framing terms, while SRA focuses on the problematic metaphor elements that human annotators identified as rationales for hate speech classification. This demonstrates SRA's effectiveness in learning to attend to subtle bias indicators beyond explicit hate terms.}
\label{fig:attention_heatmap}
\end{figure*}

\paragraph{Attention Alignment Patterns.}
Our systematic analysis reveals consistent improvements in attention-rationale alignment across rationale alignment hyperparameters, with Pearson correlation coefficients increasing from -0.084 at baseline with $\alpha=0$ to 0.649 at $\alpha=100$ for English, and similar patterns in Portuguese reaching 0.757. This suggests SRA guides model attention toward human-identified rationales.

SRA demonstrates qualitative differences from unsupervised approaches in attention distribution, as we see in this example from the English dataset: ``Go back to where you came from — but I don't hate all immigrants,''. Baseline BERT distributes attention across structural elements (highest attention on ``you'' = 0.286, ``go'' = 0.160, ``back'' = 0.120), while SRA focuses on sentiment-critical terms (``hate'' = 0.394, ``all'' = 0.445) that human annotators identified as decisive for classification. This shift indicates SRA's capacity to learn linguistically meaningful patterns beyond surface-level features.

Cross-linguistic consistency emerges in Portuguese examples. For ``Só podia ser mulher dirigindo desse jeito,'', (``It could only be a woman driving like that''), SRA concentrates attention on stereotype-indicating phrases (``jeito'' = 0.277, ``desse'' = 0.215, ``mulher'' = 0.113), while baseline models show more dispersed attention patterns, focusing on special tokens ([CLS] = 0.633). These improvements across languages indicate that SRA well aligns model attention with human rationales rather than learning dataset-specific artifacts.

\paragraph{Handling Identity Terms.}
SRA shows improved precision in aligning attention with human rationales for identity-related terminology, with token precision improving from 0.265 for unsupervised attention to 0.938 with $\alpha=10$ (see Table~\ref{tab:english_ablation} in Supplementary). This means that when SRA highlights tokens as important, they are much more likely to match the tokens that human annotators identified as rationales for their labeling decisions. BERT's WordPiece tokenization segments certain terms into subword units, allowing examination of attention patterns across these components. We analyzed attention distributions for examples containing reclaimed identity terms used in non-offensive contexts. In one case containing ``That's my nigga right there!'' (labeled as non-offensive in the dataset), BERT's tokenizer produces subword pieces [``ni'', ``\#\#gga'']. SRA allocated attention weights of 0.665 and 0.328 to these respective tokens while correctly maintaining non-offensive classification, whereas the baseline model showed more dispersed attention across function words and sentence structure. This example illustrates how rationale supervision influences attention allocation at the subword level for identity terms that undergo WordPiece segmentation. Such patterns suggest that explicit attention supervision may help models focus on relevant linguistic features when distinguishing between harmful targeting and in-group usage contexts.

\paragraph{Implicit Hate Speech.} SRA shows improved performance in detecting subtle offensive content where harmful intent is expressed indirectly. In the Portuguese example ``Esses estrangeiros vêm aqui só para atrapalhar, deveriam voltar para o país deles'' (``These foreigners come here just to cause trouble, they should go back to their country''), the model identifies xenophobic sentiment by focusing on key argumentative elements (``Esses'' = 0.126, ``deveriam'' = 0.112, ``só'' = 0.111) that construct exclusionary narratives rather than explicit slurs. As illustrated in Figure~\ref{fig:attention_heatmap}, SRA's attention patterns reveal sensitivity to rhetorical structures common in implicit offensive/hate speech. While the baseline model scatters attention across neutral framing terms, SRA focuses on problematic elements that human annotators identified as rationales. This suggests SRA learns to recognize argumentative patterns beyond lexical hate indicators, attending to subtle bias indicators such as categorical judgment terms, modal expressions of obligation, and restrictive qualifiers.

\paragraph{Performance Implications.}
Analysis of model predictions revealed potential labeling inconsistencies in the HateXplain dataset (see Supplementary for details). These discrepancies potentially impact fair model evaluation, as standard metrics may underestimate performance when models correctly identify harmful content that contradicts ground truth labels. To quantify this effect, we evaluated SRA on a filtered test set excluding those 335 identified problematic cases. We noticed that Test F1 of SRA improved from 0.682 to 0.796 (+16.7\%), accuracy from 0.696 to 0.814 (+17.0\%), and IoU F1 from 0.539 to 0.561. Fairness metrics also improved, with subgroup bias reduction showing a 9.9\% relative gain compared to baseline. While these filtered results are not directly comparable to baseline methods evaluated on the full dataset, they suggest that actual model performance may be higher than what standard benchmarks indicate, especially for fairness-critical applications. The improved performance on the filtered dataset indicates that SRA's attention supervision helps the model learn patterns that sometimes conflict with the original ground truth labels. However, we acknowledge the inherent subjectivity in hate speech annotation and the possibility that some discrepancies reflect legitimate disagreements rather than clear errors.

\subsection{Discussion}
\paragraph{Key Findings and Implications.}
The observed improvements in attention alignment metrics, from negative correlations at baseline to significantly positive correlations under a range of alignment hyperparameters, indicate that explicit rationale supervision steers models toward more human-interpretable decisions. Our findings show that models trained with SRA focus on meaningful content words, understanding identity terms in context, and recognizing subtle hate speech patterns beyond explicit slurs, while maintaining cross-linguistic applicability between English and Portuguese datasets. These improvements suggest that human reasoning can be integrated into neural architectures for sensitive classification tasks.

A limitation of our approach is that improvements in explainability come with trade-offs in fairness metrics. While SRA achieves second-best GMB-BNSP, it shows lower GMB-Subgroup AUC compared to BERT-based baselines. This suggests that rationale supervision may improve some aspects of fairness while affecting others, requiring careful evaluation in deployment scenarios. Additionally, SRA requires rationale-annotated training data, which is more expensive to obtain than standard classification labels, potentially limiting scalability to new domains or languages.

\subsection{Conclusion}
This work suggests that explicit rationale supervision offers a potentially viable path toward self-explaining hate speech detection without compromising classification performance. The SRA framework demonstrates improvements in attention alignment across languages while maintaining competitive accuracy, contributing to a broader movement toward explainable AI in sensitive domains. However, broader implications extend beyond technical metrics to fundamental questions about automated systems moderating human communication. While improved interpretability facilitates  responsible AI systems that meet desired transparency requirements, it cannot resolve the underlying tensions between protecting individuals from harm and preserving open dialogue in societies.

The systematic evaluation approach presented here provides a methodological foundation for future research at the intersection of interpretability, fairness, and performance in socially sensitive applications. As AI systems increasingly influence consequential decisions about human communication and beyond, the development of self-explaining models that better align with human rationales is an important step toward responsible AI.

\section*{Acknowledgments}
This work was supported by the Research Council of Norway through FRIPRO Grant under project number 356103, its Centres of Excellence scheme, Integreat - Norwegian Centre for knowledge-driven machine learning under project number 332645, and UiO Life Sciences Summer Internship. The computations were performed on resources provided by Educloud Research infrastructure at UiO.

\bibliography{aaai2026}

\section*{Reproducibility Checklist}

Unless specified otherwise, please answer ``yes'' to each question if the relevant information is described either in the paper itself or in a technical appendix with an explicit reference from the main paper. If you wish to explain an answer further, please do so in a section titled ``Reproducibility Checklist'' at the end of the technical appendix.

\noindent \textbf{This paper:}
\begin{itemize}
    \item Includes a conceptual outline and/or pseudocode description of AI methods introduced (yes/partial/no/NA). \textbf{yes}
    \item Clearly delineates statements that are opinions, hypothesis, and speculation from objective facts and results (yes/no). \textbf{yes}
    \item Provides well-marked pedagogical references for less-familiar readers to gain background necessary to replicate the paper (yes/no). \textbf{yes}
\end{itemize}

\noindent \textbf{Does this paper make theoretical contributions?} (yes/no) \textbf{no}

If yes, please complete the list below:
\begin{itemize}
    \item All assumptions and restrictions are stated clearly and formally (yes/partial/no).
    \item All novel claims are stated formally (e.g., in theorem statements) (yes/partial/no).
    \item Proofs of all novel claims are included (yes/partial/no).
    \item Proof sketches or intuitions are given for complex and/or novel results (yes/partial/no).
    \item Appropriate citations to theoretical tools used are given (yes/partial/no).
    \item All theoretical claims are demonstrated empirically to hold (yes/partial/no/NA).
    \item All experimental code used to eliminate or disprove claims is included (yes/no/NA).
\end{itemize}

\noindent \textbf{Does this paper rely on one or more datasets?} (yes/no) \textbf{yes}

If yes, please complete the list below:
\begin{itemize}
    \item A motivation is given for why the experiments are conducted on the selected datasets (yes/partial/no/NA). \textbf{yes}
    \item All novel datasets introduced in this paper are included in a data appendix (yes/partial/no/NA). \textbf{NA}
    \item All novel datasets introduced in this paper will be made publicly available upon publication with a license allowing free usage for research purposes (yes/partial/no/NA). \textbf{NA}
    \item All datasets drawn from the existing literature are accompanied by appropriate citations (yes/no/NA). \textbf{yes}
    \item All datasets drawn from the existing literature are publicly available (yes/partial/no/NA). \textbf{yes}
    \item All datasets that are not publicly available are described in detail with justification (yes/partial/no/NA). \textbf{NA}
\end{itemize}

\noindent \textbf{Does this paper include computational experiments?} (yes/no) \textbf{yes}

If yes, please complete the list below:
\begin{itemize}
    \item This paper states the number and range of values tried per (hyper-)parameter and the criterion used for selecting the final setting (yes/partial/no/NA). \textbf{yes}
    \item Any code required for preprocessing data is included in the appendix (yes/partial/no). \textbf{yes}
    \item All source code required for conducting and analyzing the experiments is included in a code appendix (yes/partial/no). \textbf{no. The code, including rationale mask construction,
will be released upon publication.}
    \item All source code will be made publicly available upon publication with an open research license (yes/partial/no). \textbf{yes}
    \item All source code implementing new methods includes comments with references to the paper (yes/partial/no). \textbf{yes}
    \item If randomness is used, seed-setting is described sufficiently for replication (yes/partial/no/NA). \textbf{yes}
    \item The computing infrastructure is specified (hardware, OS, libraries, versions) (yes/partial/no). \textbf{yes}
    \item Evaluation metrics are formally described and justified (yes/partial/no). \textbf{yes}
    \item The number of algorithm runs used for each result is stated (yes/no). \textbf{yes}
    \item Results include variation/confidence measures beyond simple averages (yes/no). \textbf{yes}
    \item The significance of performance differences is judged with statistical tests (yes/partial/no). \textbf{no. We reported error bars with multiple runs for key experiments.}
    \item All final (hyper-)parameters used are listed (yes/partial/no/NA). \textbf{yes}
\end{itemize}

\newpage
\appendix
\onecolumn

\begin{center}\Huge Supplementary Material
\end{center}

\section{Complete Related Work}
Hate speech explainable methods are commonly categorized into two aspects: \textit{self-explaining} or \textit{post-hoc explaining} \cite{10.1145/3236009}, which may provide local our global explanations. Local explanations are provided for individual instances, while global explanations apply to the model's behavior across any input \cite{balkir-etal-2022-challenges}. Self-explaining methods rely on the internal structure of the prediction model, making these methods often tailored to specific model types. In contrast, post-hoc explaining  methods do not rely on knowledge of the to-be-explained model, but only input-output pairs \cite{balkir-etal-2022-challenges}. %

\subsection{Self-Explaining Hate Speech Detection}

\citet{mathew2021hatexplain} proposed baseline models for explainable hate speech detection with human rationales. They used human-annotated rationales, which are spans of text highlighted by annotators as justification for the hate/offensive/normal labels. They transform these rationales into ground truth attention vectors (e.g., for each token in a post, a binary vector indicates whether it was part of the highlighted rationale (1) or not (0)). Thus, they average across multiple annotators and apply a softmax with temperature to create a probability distribution over tokens (ground truth attention). During training, for attention-based models (BiRNN-Attention and BERT), they computed a cross-entropy loss between the model’s predicted attention weights and the ground truth attention vectors. This encourages the model’s attention mechanism to align with human-provided rationales. The models are then evaluated not only on standard performance metrics (accuracy, macro-F1, AUROC), but also on explainability metrics such as  plausibility (IoU-F1, Token-F1, AUPRC) and faithfulness (comprehensiveness, sufficiency), following the ERASER benchmark \cite{deyoung-etal-2020-eraser}. Finally, in terms of interpretability, the BiRNN-HateXplain [Attn] model provided the most plausible and faithful explanations.

\citet{kim-etal-2022-hate} proposed the Masked Rationale Prediction (MRP), which consists of an intermediate token-level classification task. Given an input sentence with rationale labels (1 for rationale tokens, 0 otherwise), a portion of the rationale embeddings is masked (replaced with zero vectors). The model receives the input token embeddings plus these partially masked rationale embeddings and is trained to predict the masked rationale labels. Formally, the input is given by Equation \ref{eq1}.

\begin{equation}
H^{(0)}_{\mathrm{MRP}} = X_S + \tilde{X}_R
\label{eq1}
\end{equation}

where $H^{(0)}_{\mathrm{MRP}}$] consists of initial hidden state of the MRP, [$X_S$] embeddings of the sentence tokens, and [$\tilde{X}_R$] embeddings of the rationales, partially masked. The model minimizes cross-entropy loss only on the masked positions. After this, the BERT parameters are fine-tuned on the downstream hate speech classification task, leveraging the context-aware reasoning learned during MRP.

\citet{calabrese-etal-2022-explainable} proposed a self-explaining approach for abuse detection formulated as an Intent Classification and Slot Filling (ICSF) task. It employs a two-stage sequence-to-sequence model based on BART: first, it generates a meaning sketch identifying relevant slots (e.g., Target, ProtectedCharacteristic); then, it refines this sketch by filling in the corresponding textual spans. The intent (e.g., Dehumanization, Derogation) is deterministically inferred from the filled slots, ensuring that the model’s prediction is directly traceable to the extracted evidence. This structured output serves as an intrinsic, human-readable explanation of the classification decision, rather than a post-hoc justification.

\citet{clarke-etal-2023-rule} introduce \textit{Rule By Example} (RBE), an exemplar-based contrastive learning framework for explainable hate-speech detection. RBE jointly trains a dual-encoder, a text encoder $\Theta_t$ and a rule encoder $\Theta_r$, to align inputs with human-authored logical rules and their exemplars. For each input $x_t$, applicable rule exemplars $e$ are concatenated and encoded, so learning minimizes a contrastive loss over the cosine distance $D$ between the rule and text embeddings:
\[
\mathcal{L}=\tfrac{1}{2}\big(Y D^2 + (1-Y)\,\max(m-D,0)^2\big),
\]
where $Y\in\{0,1\}$ indicates whether a rule governs $x_t$ and $m$ is a margin. This alignment enables \emph{rule-grounding}: predictions are explained by the most similar rules and their nearest exemplars, yielding transparent and easily customizable moderation behavior. RBE is evaluated in supervised and unsupervised settings on HateXplain, Jigsaw, and CAD; on HateXplain it reports macro-F1 $\approx 0.816$ using BERT with the HateXplain ruleset, while providing faithful traceability to human-readable rules rather than token-level rationales.

\citet{nirmal-etal-2024-towards} presented SHIELD, a rationale-augmented framework for hate speech detection that leverages GPT-3.5 to extract textual rationales via prompt-based querying. The input text is encoded using HateBERT, while the extracted rationales are embedded using a frozen BERT (bert-base-uncased). The [CLS] embeddings from both encoders are concatenated and passed through a two-layer MLP with ReLU activation. The model is trained using binary cross-entropy loss, promoting faithful interpretability by design through the inclusion of rationale-derived representations.

\subsection{Post-Hoc Explaining Hate Speech Detection}

\citet{mosca-etal-2023-ifan} introduced IFAN is a framework for real-time, human-in-the-loop interaction with NLP models via post-hoc explanations (LIME) and adapter-based fine-tuning. Tested on HateXplain (hate speech detection) and GYAFC (formality classification) datasets, models like BERT and BLOOM were used. Feedback on explanations is integrated via adapter layers, improving model debiasing—e.g., BERT precision on the Jewish subgroup improved from 0.95 to 0.97 with balanced feedback. LLMs like GPT-4X Alpaca reached F1=0.64. IFAN offers APIs, user roles, and visualization tools to enhance model transparency and fairness.

\citet{yang-etal-2023-hare} introduced HARE, a framework for hate speech detection using step-by-step reasoning generated by LLMs via Chain-of-Thought (CoT) prompting. The authors proposed two methods: (i) Fr-HARE, which generates rationales from scratch without human annotations, and (ii) Co-HARE, which generates rationales conditioned on human-provided annotations. Results On SBIC \cite{sap-etal-2020-social} and Implicit Hate \cite{elsherief-etal-2021-latent} dataset benchmarks, Co-HARE achieved 85.35\% accuracy and 85.93\% F1, surpassing the human-annotation baseline (84.23\% acc / 85.21\% F1). In addition, HARE improves generalization to unseen datasets such as HateXplain and DynaHate.

\citet{wasi-2024-explainable} presented XG-HSI, a Graph Neural Network (GNN)-based framework for the explainable Islamophobic hate speech. A subset of the HateXplain dataset focused on Muslims was used. The method converts posts into graphs where each node represents a post, and edges connect posts based on contextual similarity (cosine similarity on BERT embeddings). The GNN architecture uses attention mechanisms and semi-supervised learning for feature extraction and classification, as well as GNNExplainer to provide insights into the model's predictions by identifying influential nodes (tokens) and connections. The XG-HSI-BERT model outperformed the current baseline (F1 of 0.747).

\citet{salles-etal-2025-hatebrxplain} employed two model-agnostic post-hoc explanation methods: LIME (Local Interpretable Model-agnostic Explanations) and SHAP (SHapley Additive exPlanations). These methods were applied to generate token-level rationales over a 10\% sample of the offensive instances from the HateBRXplain dataset in Portuguese. Results showed that while post-hoc explanations by SHAP showed high alignment with human annotations, the best-performing classification model (BERTimbau) did not yield the most faithful explanations, highlighting a trade-off between predictive performance and explainability.

\citet{dementieva-etal-2025-multilingual} introduced the multilingual detoxification dataset (9 languages) and evaluates models such as Delete, condBERT, mBART, and GPT-4 CoT, which clusters inputs by toxicity features and applies tailored Chain-of-Thought prompting. GPT-4 CoT achieved the best average joint score (0.331), e.g., 0.447 (German), 0.503 (Ukrainian), outperforming mBART (0.282) and Delete (0.302), with improved precision and cultural sensitivity.

\section{Additional Experiments}
\begin{table*}[!htbp]
\centering
\caption{Impact of the rationale alignment hyperparameter ($\alpha$) on BERT performance for English hate speech detection. Higher $\alpha$ values indicate stronger supervision.}
\label{tab:english_ablation}
\begin{tabular}{c|c|c|c|c|c|c}
\hline
$\alpha$ & Test F1 ↑ & Accuracy ↑ & Attn-Rationale ↑ & IoU F1 ↑ & Token Prec ↑ & Token F1 ↑ \\
\hline
0.0 (baseline) & 0.671 & 0.684 & -0.084 & 0.019 & 0.265 & 0.122 \\
0.05 & 0.670 & 0.679 & 0.475 & 0.397 & 0.831 & 0.540 \\
0.1 & 0.687 & 0.699 & 0.504 & 0.398 & 0.866 & 0.545 \\
0.2 & 0.675 & 0.686 & 0.534 & 0.447 & 0.886 & 0.574 \\
0.5 & 0.669 & 0.674 & 0.573 & 0.477 & 0.913 & 0.605 \\
1.0 & 0.685 & 0.689 & 0.587 & 0.503 & 0.919 & 0.619 \\
2.0 & 0.679 & 0.696 & 0.602 & 0.517 & 0.928 & 0.629 \\
5.0 & 0.677 & 0.689 & 0.616 & 0.529 & 0.935 & 0.640 \\
10.0 & 0.683 & 0.695 & 0.624 & 0.537 & 0.938 & 0.648 \\
20.0 & 0.676 & 0.688 & 0.635 & 0.550 & 0.944 & 0.662 \\
100.0 & 0.694 & 0.700 & 0.649 & 0.572 & 0.942 & 0.678 \\
\hline
\end{tabular}
\end{table*}

\begin{table*}[!htbp]
\centering
\caption{Impact of the rationale alignment hyperparameter ($\alpha$) on BERTimbau performance for Portuguese hate speech detection.}
\label{tab:portuguese_ablation}
\begin{tabular}{c|c|c|c|c|c|c}
\hline
$\alpha$ & Test F1 ↑ & Accuracy ↑ & Attn-Rationale ↑ & IoU F1 ↑ & Token Prec ↑ & Token F1 ↑ \\
\hline
0.0 (baseline) & 0.903 & 0.903 & -0.385 & 0.000 & 0.000 & 0.000 \\
0.05 & 0.916 & 0.914 & 0.557 & 0.387 & 0.918 & 0.574 \\
0.1 & 0.910 & 0.907 & 0.555 & 0.419 & 0.917 & 0.584 \\
0.2 & 0.923 & 0.920 & 0.628 & 0.493 & 0.937 & 0.625 \\
0.5 & 0.921 & 0.919 & 0.646 & 0.521 & 0.946 & 0.652 \\
1.0 & 0.916 & 0.913 & 0.680 & 0.611 & 0.956 & 0.693 \\
2.0 & 0.915 & 0.913 & 0.698 & 0.622 & 0.949 & 0.707 \\
5.0 & 0.917 & 0.914 & 0.723 & 0.682 & 0.939 & 0.746 \\
10.0 & 0.906 & 0.904 & 0.725 & 0.691 & 0.939 & 0.750 \\
20.0 & 0.902 & 0.900 & 0.746 & 0.710 & 0.925 & 0.764 \\
100.0 & 0.905 & 0.904 & 0.757 & 0.751 & 0.913 & 0.771 \\
\hline
\end{tabular}
\end{table*}

\begin{table*}[!htbp]
\centering
\caption{Ablation study: SRA performance by attention layer (averaged over all heads) on HateXplain (English). Results show that SRA is robust to the choice of attention layer.}
\label{tab:ablation_layers}
\begin{tabular}{lcccc}
\toprule
\textbf{Config} & \textbf{Test F1} & \textbf{IoU F1} & \textbf{Token F1} & \textbf{AUPRC} \\
\midrule
Layer 6          & 0.6835 & 0.526 & 0.641 & 0.746 \\
Layer 7          & 0.6722 & 0.523 & 0.639 & 0.746 \\
Layer 8          & 0.6871 & 0.533 & 0.648 & 0.751 \\
Layer 9          & 0.6713 & 0.541 & 0.650 & 0.751 \\
Layer 10         & 0.6809 & 0.531 & 0.647 & 0.753 \\
Layer 11 (last)  & 0.6825 & 0.537 & 0.649 & 0.752 \\
\midrule
\textit{Std}     & 0.006 & 0.006 & 0.004 & 0.003 \\
\bottomrule
\end{tabular}
\end{table*}

\begin{table*}[!htbp]
\centering
\caption{Head-wise ablation study on BERT layer 8, HateXplain (English). Results for all 12 heads (0--11) are shown.}
\label{tab:head_ablation_layer8}
\begin{tabular}{lcccc}
\toprule
\textbf{Config} & \textbf{Test F1} & \textbf{IoU F1} & \textbf{Token F1} & \textbf{AUPRC} \\
\midrule
layer 8 head 0  & 0.6762 & 0.541 & 0.651 & 0.755 \\
layer 8 head 1  & 0.6650 & 0.546 & 0.656 & 0.759 \\
layer 8 head 2  & 0.6763 & 0.540 & 0.652 & 0.754 \\
layer 8 head 3  & 0.6739 & 0.541 & 0.650 & 0.753 \\
layer 8 head 4  & 0.6745 & 0.537 & 0.651 & 0.752 \\
layer 8 head 5  & 0.6853 & 0.551 & 0.656 & 0.757 \\
layer 8 head 6  & 0.6853 & 0.551 & 0.656 & 0.757 \\
layer 8 head 7  & 0.6826 & 0.537 & 0.648 & 0.753 \\
layer 8 head 8  & 0.6831 & 0.543 & 0.657 & 0.757 \\
layer 8 head 9  & 0.6804 & 0.539 & 0.651 & 0.758 \\
layer 8 head 10 & 0.6818 & 0.541 & 0.656 & 0.756 \\
layer 8 head 11 & 0.6773 & 0.528 & 0.648 & 0.755 \\
\midrule
\textit{Std} & 0.006 & 0.006 & 0.003 & 0.002 \\
\bottomrule
\end{tabular}
\end{table*}

\begin{table*}[!htbp]
\centering
\caption{Bias Ablation: Fairness metrics (GMB-Subgroup, GMB-BPSN, and GMB-BNSP AUC) across different values of rationale supervision weight $\alpha$ on HateXplain (English).}
\label{tab:bias_ablation}
\resizebox{\linewidth}{!}{
\begin{tabular}{lccccccccccc}
\toprule
\textbf{Metric} & \textbf{0} & \textbf{0.05} & \textbf{0.1} & \textbf{0.2} & \textbf{0.5} & \textbf{1.0} & \textbf{2.0} & \textbf{5.0} & \textbf{10.0} & \textbf{20.0} & \textbf{100.0} \\
\midrule
\textbf{GMB-Sub}  & 0.7165 & 0.7143 & 0.7220 & 0.7212 & 0.7128 & 0.7171 & 0.7161 & 0.7156 & 0.7120 & 0.7132 & 0.7157 \\
\textbf{GMB-BPSN} & 0.7170 & 0.7191 & 0.7174 & 0.7200 & 0.7192 & 0.7197 & 0.7194 & 0.7195 & 0.7211 & 0.7186 & 0.7192 \\
\textbf{GMB-BNSP} & 0.8169 & 0.8212 & 0.8294 & 0.8318 & 0.8244 & 0.8179 & 0.8254 & 0.8168 & 0.8186 & 0.8208 & 0.8204 \\
\bottomrule
\end{tabular}}
\end{table*}

\begin{table*}[!htbp]
\centering
\caption{Bias Ablation (Portuguese): Fairness metrics (GMB-Subgroup, GMB-BPSN, and GMB-BNSP AUC) across different values of rationale supervision weight $\alpha$ on HateBRXplain (Portuguese).}
\label{tab:bias_ablation_pt}
\resizebox{\linewidth}{!}{
\begin{tabular}{lccccccccccc}
\toprule
\textbf{Metric} & \textbf{0} & \textbf{0.05} & \textbf{0.1} & \textbf{0.2} & \textbf{0.5} & \textbf{1.0} & \textbf{2.0} & \textbf{5.0} & \textbf{10.0} & \textbf{20.0} & \textbf{100.0} \\
\midrule
\textbf{GMB-Sub}  & 0.9011 & 0.9197 & 0.9252 & 0.9057 & 0.8954 & 0.8680 & 0.8956 & 0.8773 & 0.8818 & 0.8687 & 0.8464 \\
\textbf{GMB-BPSN} & 0.9343 & 0.9314 & 0.9420 & 0.9195 & 0.9150 & 0.8875 & 0.8962 & 0.8919 & 0.9227 & 0.8992 & 0.9019 \\
\textbf{GMB-BNSP} & 0.9345 & 0.9630 & 0.9683 & 0.9655 & 0.9586 & 0.9616 & 0.9595 & 0.9328 & 0.9304 & 0.9308 & 0.9214 \\
\bottomrule
\end{tabular}}
\end{table*}

\newpage 

\section{Expert Review and Dataset Considerations}

We discovered potential labeling inconsistencies in the HateXplain dataset. In the test sample, 597 out of 1,922 classifications (31.1\%) made by SRA appeared to be erroneous. Through manual inspection of the labels and texts we found that many of the texts seemed to represent instances where our model correctly identified content characteristics despite disagreement with original labels.
Two NLP researchers with backgrounds in hate speech detection independently reviewed the texts thought to be misclassified according to the error analysis. They identified 335 problematic cases (56.1\% of disagreements) with potential systematic patterns, with instances labeled ``Normal'' containing derogatory identity-targeted language, content marked ``Hate speech'' lacking clear targeting criteria, and inconsistent offensive/hate speech boundaries. These human misclassification patterns may reflect inherent challenges in crowdsourced annotation frameworks. HateXplain employed multiple annotators per instance, but the complexity of hate speech categorization can lead to annotation variability even among trained raters. It can be especially difficult to distinguish hate speech from offensive examples. Prior work has documented similar challenges in large-scale annotation projects for subjective tasks \cite{Davidson_Warmsley_Macy_Weber_2017,snow2008cheap}.

\section{Broader societal impact and ethical considerations}

The development of more interpretable hate speech detection systems carries complex implications that extend beyond technical performance metrics. SRA could help address legal transparency requirements such as GDPR's right to meaningful information about automated decisions. Yet the deployment of such systems raises fundamental questions about freedom of expression and democratic discourse.

One potential risk is that hate speech detection systems could disproportionately affect the communities they are designed to protect. Individuals from marginalized backgrounds may use language patterns that automated systems flag as offensive, despite the absence of harmful intent. This presents a risk where protective measures could inadvertently silence the voices of those most vulnerable to discrimination. Our approach of incorporating human rationales addresses this problem by making value judgments more transparent, but it cannot resolve the underlying tension between different perspectives on harmful speech. It is also likely that as detection methods become more sophisticated, users will circumvent moderation by developing increasingly subtle ways to express harmful content while avoiding detection. This dynamic suggests that technological solutions alone cannot address the complex social phenomenon of online hate speech.

The implications of rationale-supervised attention extend beyond hate speech detection to other high-stakes domains where interpretability is crucial. The approach could prove valuable in criminal justice for risk assessment algorithms, healthcare for diagnostic systems, education for performance prediction, employment for hiring algorithms, and finance for credit scoring. The common thread across these applications is the need for both accuracy and accountability in decisions that significantly affect people's lives.

\end{document}